\documentclass{ieeeaccess}
\usepackage{cite}
\usepackage{amsmath,amssymb,amsfonts}
\usepackage{algorithmic}
\usepackage{graphicx}
\usepackage{textcomp}
\usepackage{kotex}
\usepackage{hyperref}

\usepackage{bm}
\makeatletter
\AtBeginDocument{\DeclareMathVersion{bold}
\SetSymbolFont{operators}{bold}{T1}{times}{b}{n}
\SetSymbolFont{NewLetters}{bold}{T1}{times}{b}{it}
\SetMathAlphabet{\mathrm}{bold}{T1}{times}{b}{n}
\SetMathAlphabet{\mathit}{bold}{T1}{times}{b}{it}
\SetMathAlphabet{\mathbf}{bold}{T1}{times}{b}{n}
\SetMathAlphabet{\mathtt}{bold}{OT1}{pcr}{b}{n}
\SetSymbolFont{symbols}{bold}{OMS}{cmsy}{b}{n}
\renewcommand\boldmath{\@nomath\boldmath\mathversion{bold}}}

\makeatother

\def\BibTeX{{\rm B\kern-.05em{\sc i\kern-.025em b}\kern-.08em
    T\kern-.1667em\lower.7ex\hbox{E}\kern-.125emX}}

\begin{document}
\history{Date of publication xxxx 00, 0000, date of current version xxxx 00, 0000.}
\doi{10.1109/ACCESS.2024.0429000}

\title{EndoForce: Development of an Intuitive Axial Force Measurement Device for Endoscopic Robotic Systems}
\author{\uppercase{Hansoul Kim}\authorrefmark{1}, \uppercase{Dong-Ho Lee}\authorrefmark{2}, \uppercase{Dukyoo Kong}\authorrefmark{2}, \uppercase{Dong-Soo Kwon}\authorrefmark{2, 3},  \MakeUppercase{Byungsik Cheon}\authorrefmark{4}}

\address[1]{Department of Mechanical Engineering, Myongji University, Yongin-si, Gyeonggi-do, Republic of Korea, 17058 (e-mail: hansoul@mju.ac.kr)}
\address[2]{R\&D Department, Roen Surgical, Inc., Daejeon, Republic of Korea (e-mail: \{leedh, kongdy, kwonds\}@roensurgical.com)}
\address[3]{Department of Mechnical Engineering, Korea Advanced Institute of Science and Technology, Daejeon, Republic of Korea (e-mail: kwonds@kaist.ac.kr)}
\address[4]{School of Mechatronics Engineering, Korea University of Technology and Education, 1600 Chungjeol-ro, Dongnam-gu, Cheonan-si, Chungnam, Republic of Korea, 31253 (e-mail: cbs@koreatech.ac.kr)}

\tfootnote{This work was supported by 2024 Research Fund of Myongji University.}

\markboth
{Kim \headeretal: EndoForce: Development of an Intuitive Axial Force Measurement Device for Endoscopic Robotic Systems}
{Kim \headeretal: EndoForce: Development of an Intuitive Axial Force Measurement Device for Endoscopic Robotic Systems}

\corresp{Corresponding author: Byungsik Cheon (e-mail: cbs@koreatech.ac.kr).}

\begin{abstract}
Robotic endoscopic systems provide intuitive control and eliminate radiation exposure, making them a promising alternative to conventional methods. However, the lack of axial force measurement from the robot remains a major challenge, as it can lead to excessive colonic elongation, perforation, or ureteral complications. Although various methods have been proposed in previous studies, limitations such as model dependency, bulkiness, and environmental sensitivity remain challenges that should be addressed before clinical application. In this study, we propose EndoForce, a device designed for intuitive and accurate axial force measurement in endoscopic robotic systems. Inspired by the insertion motion performed by medical doctors during ureteroscopy and gastrointestinal (GI) endoscopy, EndoForce ensures precise force measuring while maintaining compatibility with clinical environments. The device features a streamlined design, allowing for the easy attachment and detachment of a sterile cover, and incorporates a commercial load cell to enhance cost-effectiveness and facilitate practical implementation in real medical applications. To validate the effectiveness of the proposed EndoForce, physical experiments were performed using a testbed that simulates the ureter. We show that the axial force generated during insertion was measured with high accuracy, regardless of whether the pathway was straight or curved, in a testbed simulating the human ureter.
\end{abstract}

\begin{keywords}
Endoscopic robotic systems, Force sensing, Haptic feedback, Ureteroscopy, Gastrointestinal endoscopy.
\end{keywords}

\titlepgskip=-21pt

\maketitle
\section{Introduction}\label{sec:introduction}
\begin{figure*}[t]
    \centering
    \includegraphics[width=0.9\textwidth]{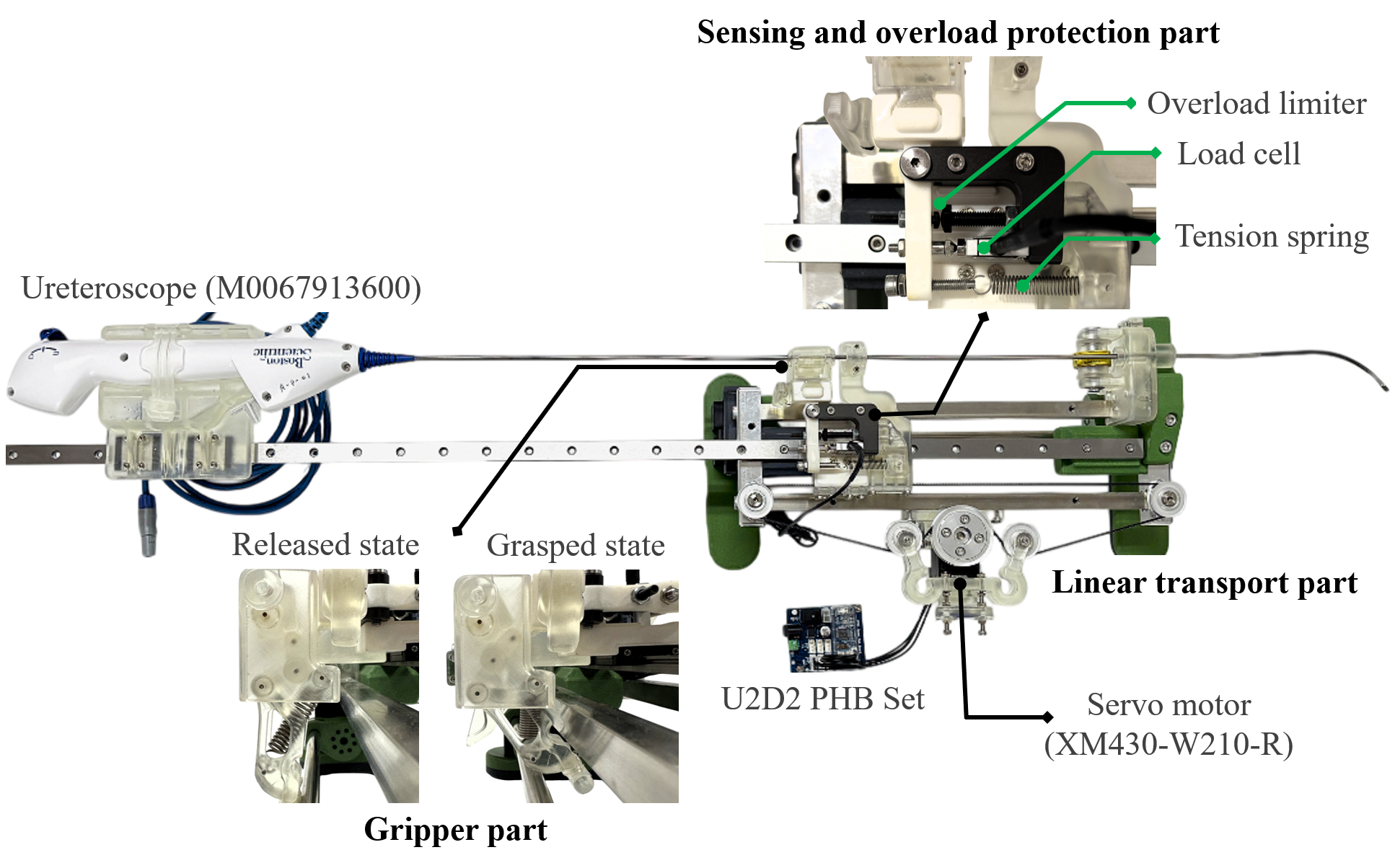}
    \caption{An overview of the proposed force feedback device, EndoForce, capable of measuring the axial force from the endoscopic robotic system. The proposed device consists of three main components: a sensing part to measure the axial force; an overload protection part that prevents overload on the load cell; a gripper part for the stable insertion of the insertion tube; and a linear transport part to automate the forward and backward movements of the scope or the endoscopic robotic system.}
    \label{fig:overview}
\end{figure*}

Gastrointestinal (GI) cancer and urothelial carcinoma are among the major cancer types worldwide, with high incidence and mortality rates \cite{sung2021global}. These types of cancer typically begin as polyps or non-invasive lesions, which gradually progress through years of carcinogenesis to eventually develop into malignancies. However, early detection and appropriate treatment can significantly improve survival rates. For GI cancer, the 5-year survival rate is reported to be high when diagnosed at an early stage \cite{japanese2023japanese, smith2002american}, and for urothelial carcinoma, early detection allows for bladder-preserving treatments and increases the chances of prolonged survival \cite{kamat2016definitions}.

Endoscopy or ureteroscopy is one of the most representative methods for screening,  early detection, and treating diseases such as GI cancer, urothelial carcinoma, and upper urinary tract disorders like kidney stones \cite{hamashima2015mortality, inoue2021retrograde}. However, conventional endoscopy and ureteroscopy have the following disadvantages:
\begin{itemize}
    \item Both methods, with their heavy weight and non-intuitive manipulation, impose physical and mental burdens even on highly experienced medical doctors, making it difficult to perform the procedure multiple times a day  \cite{lee2010two, tjiam2014ergonomics}.
    \item In particular, during ureteroscopy, medical doctors are exposed to the risk of X-ray radiation throughout the procedure \cite{boeri2020impact}.
\end{itemize}

To overcome the aforementioned drawbacks, various endoscopic robotic systems have been developed and are gaining attention as next-generation surgical robot platforms \cite{berthet20182, nageotte2020stras, nakadate2020surgical, hwang2020k, kim2023endoscopic, desai2011robotic , saglam2014new, rassweiler2018robot, zhao2021design}. Since the operator remotely controls the entire robotic system using an ergonomically designed master device, it not only provides more intuitive control but also eliminates the risk of radiation exposure. Moreover, the robot’s enhanced dexterity further increases its potential to improve outcomes in endoscopic surgery. However, there are still technical limitations that need to be addressed before robotic systems can be used in clinical environments.

In contrast to laparoscopic surgery, which uses a rigid robotic platform, endoscopic surgery requires the flexible robot to insert through the natural orifices and navigate tortuous anatomical pathways, such as the GI tract or urinary tract, to reach the target lesion \cite{valdastri2012advanced, huynh2017retained}. In conventional endoscopic surgery, medical doctors manipulate the scope while relying on force feedback, employing skilled insertion techniques developed through clinical experience to navigate to the target lesion \cite{lee2010two, de2020complications}. However, most endoscopic robotic systems that are remotely controlled using a master device have limitations in accurately receiving feedback on the forces generated from the robot, which leads to the following clinical issues:

\begin{itemize}
    \item The large intestine can be classified into fixed and non-fixed sites, with the non-fixed sites requiring the medical doctor to use skilled insertion techniques to pass through. In the absence of force feedback, the limitations of these insertion techniques can result in excessive elongation or colon perforation, leading to pain and complications for the patient \cite{polter2015risk, kim2021sigmoid}.
    \item Lack of force feedback in robotic ureteroscopy increases the risk of procedural complications. In particular, when the bending lever remains engaged during retraction, the endoscope tip stays in a curved position. In such cases, the scope may scrape against the ureteral wall or renal tissue while navigating through tortuous paths or even along straight segments. This unintended contact can lead to tissue damage, bleeding, or perforation, posing significant risks to patient safety \cite{thakur2020trapped, huynh2017retained, gadzhiev2017valve}.
\end{itemize}
Therefore, force feedback is one of the essential technologies that should not be absent in endoscopic robotic systems. Force feedback allows the medical doctor to monitor the applied force in real time, enabling more precise manipulation and supporting safe surgery even within complex anatomical structures \cite{wang2019novel}.

\begin{figure*}[t]
    \centering
    \includegraphics[width=1.0\textwidth]{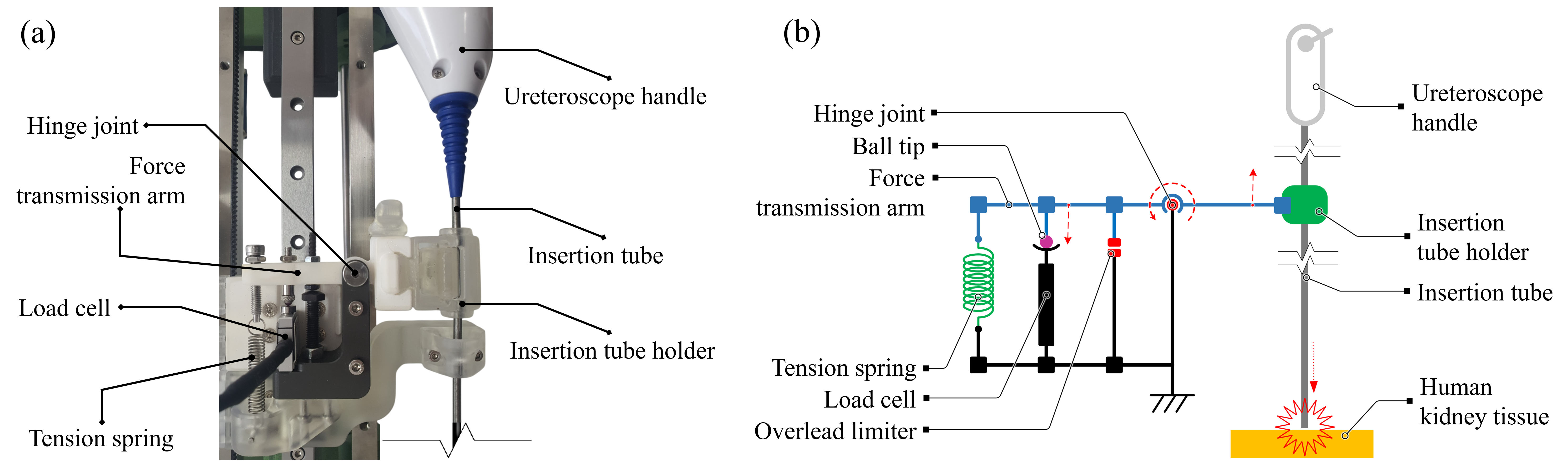}
    \caption{(a) Structural components of the sensing part, including the insertion tube holder, force transmission arm, hinge joint, ball tip, load cell, tension spring, and overload limiter; and (b) a simplified kinematic diagram illustrating the operating principle.}
    \label{fig:sensing}
\end{figure*}

Various studies have been performed to obtain force feedback information from the robotic systems. Studies can be broadly classified into two methods: predicting force feedback through modeling or deep learning \cite{li2019distal, kang2020learning} and directly measuring it by attaching additional sensors \cite{mckinley2015single, srivastava2016design, arabagi2015biocompatible, kim2017surgical, ju2018variable, presti2020fiber, lai2019force, lai2021three}. Li et al. proposed a deep learning approach to predict the distal force of tendon-sheath mechanisms using proximal end measurements \cite{li2019distal}. Kang et al. proposed a learning-based method to predict distal end forces in tendon-sheath-driven soft wearable hand robots by integrating dynamic information from motor encoders, wire tension, and sheath bending angles \cite{kang2020learning}. Generally, these studies indirectly predict the distal end force by using or combining information measured at the proximal end, making them highly dependent on predefined models and limiting their ability to adapt to changing environmental factors in real time. Unlike studies that predict force indirectly, research has been proposed that uses sensors designed based on technologies such as displacement, resistance, pressure, capacitance, and piezoelectricity to directly measure distal end force \cite{mckinley2015single, srivastava2016design, arabagi2015biocompatible, kim2017surgical, ju2018variable}. However, most existing sensors are too bulky for integration into endoscopic robotic systems, using additional electrical sensors, aside from the endoscopic camera, is not clinically appropriate. Recently, research has also been proposed that applies Fiber Bragg Gratings (FBGs) sensors, which offer various advantages such as compact size, high sensitivity, and electrical passivity \cite{presti2020fiber, lai2019force, lai2021three}. Despite these advantages, FBG sensors have limitations such as sensitivity to temperature, complex design, and high costs, making them challenging to apply in clinical environments. Therefore, there is a need for a method that is clinically appropriate, highly accurate, cost-effective, and easily compatible with endoscopic robotic systems.

Inspired by the actual insertion motion of medical doctors in conventional endoscopy and ureteroscopy, we propose EndoForce, a new device that can measure the axial force from the endoscopic robotic system (Fig. \ref{fig:overview}). The main contribution of this study can be summarized as follows:

\begin{itemize}
    \item By simulating the way medical doctors perceive force during insertion, the device can measure the same magnitude of axial force and replicate the force transmission pattern. This provides a familiar handling experience for users accustomed to manual endoscopy.
    
    \item The proposed device is designed for easy attachment and detachment of the sterile cover, ensuring compatibility with real clinical environments.

    \item The use of only commercial load cell makes the proposed device cost-effective, with simplified design and application.
\end{itemize}

% By simulating the manner in which medical doctors feel the force when they manually grip and insert the insertion tube, this device enables more intuitive and accurate measure of the distal end force. In addition, the proposed device is designed to allow easy attachment and detachment of the sterile cover, considering its use in actual clinical environments. 

% Considering the sterilization issues in actual clinical environments, it is designed with a detachable mechanism.

% The main contribution of this can be summarized as follows:
% \begin{itemize}
%     \item Contribution 1.
%     \item Contribution 2.
% \end{itemize}

The remainder of this paper is organized as follows: Section \ref{sec:method} describes the requirements, mechanical design, and control circuit of the proposed force feedback device. Sections \ref{sec:experimental_design} and \ref{sec:results} describe the physical experimental setup and the results. Sections \ref{sec:discussion} and \ref{sec:conclusion} present the discussion and conclusions of this study, along with an outline for future work.

\section{Development of Force Feedback Device: EndoForce}\label{sec:method}
This study proposes EndoForce, a new device for measuring the axial force from an endoscopic robotic system. The proposed device was designed with the following requirements, considering its application in real clinical environments.

\subsection{Requirements}\label{sec:method_requirements}
\subsubsection{Reliability and Cost Effectiveness}\label{sec:method_requirements_cost} 
The reliability of the sensor is essential for ensuring consistent clinical performance of the device, while cost-effectiveness should be considered to improve its applicability as a commercial product. 

\subsubsection{Safety}\label{sec:method_requirements_safety} 
In real clinical environments, unexpected overloads such as excessive compression or tension may occur, potentially damaging the sensors embedded in the proposed device. If the device is compromised due to overload, accurate force feedback becomes difficult, which may ultimately pose a risk to patient safety. Therefore, to ensure clinical safety, a protection mechanism that prevents device damage caused by overload must be considered.

\subsubsection{Sterilization}\label{sec:method_requirements_sterilization} 
Since the insertion tube of the endoscopic robot directly contacts the patient's biological tissue, the mechanism holding it may pose a risk of contamination. Therefore, the mechanical design for holding the insertion tube should account for the possibility of making it replaceable with detachable and disposable components.

\subsection{Mechanical Design}\label{sec:method_mechanical_design}
The overall mechanical structure consists of sensing part to measure the axial force, an overload protection part that prevents overload on the load cell, a gripper part for stable insertion of the insertion tube, and a linear transport part that automates the forward and backward movements of the endoscopic robotic system (Fig. \ref{fig:overview}).

\subsubsection{Sensing Part}\label{sec:method_md_sensor} 
Fig. \ref{fig:sensing}(a) illustrates the overall structural configuration of the sensing part. The sensing part primarily consists of the following components: insertion tube holder, force transmission arm, hinge joint, ball tip, load cell (UU74-K2, DACELL), tension spring, and overload limiter. 

Fig. \ref{fig:sensing}(b) is a simplified kinematic diagram that provides a detailed explanation of the operating principle of the sensing part. The insertion tube holder secures and guides the insertion tube while simultaneously transmitting the axial force generated at the end of the endoscopic robotic system or scope to the force transmission arm. The force transmission arm rotates around the hinge joint, reversing the direction of the force transmitted from the insertion tube holder and acting as a lever to deliver the force to the load cell. Equipped with a ball tip, tension spring, and overload limiter, it ensures stable and accurate force transmission. The hinge joint serves as the rotational axis of the force transmission arm, enabling the ruby ball tip to be preloaded by the tension spring. This configuration ensures stable contact between the ball tip and the seat while reversing the force direction transmitted from the insertion tube holder. The ball tip, positioned on the force transmission arm, forms a point contact with the ball contact seat, minimizing friction and transmitting only the axial force to the load cell. The tension spring connected to the arm applies a preload that ensures stable contact between the ball tip and the seat. This preload prevents nonlinearity near the zero region during force sensing by the load cell, allowing forces to be measured accurately within the linear response range. Finally, the load cell mounted on the sensing part converts the transmitted force into an electrical signal with high precision and functions as a key component of the system.

\begin{figure}[t]
   \centering
   \includegraphics[width=0.5\textwidth]{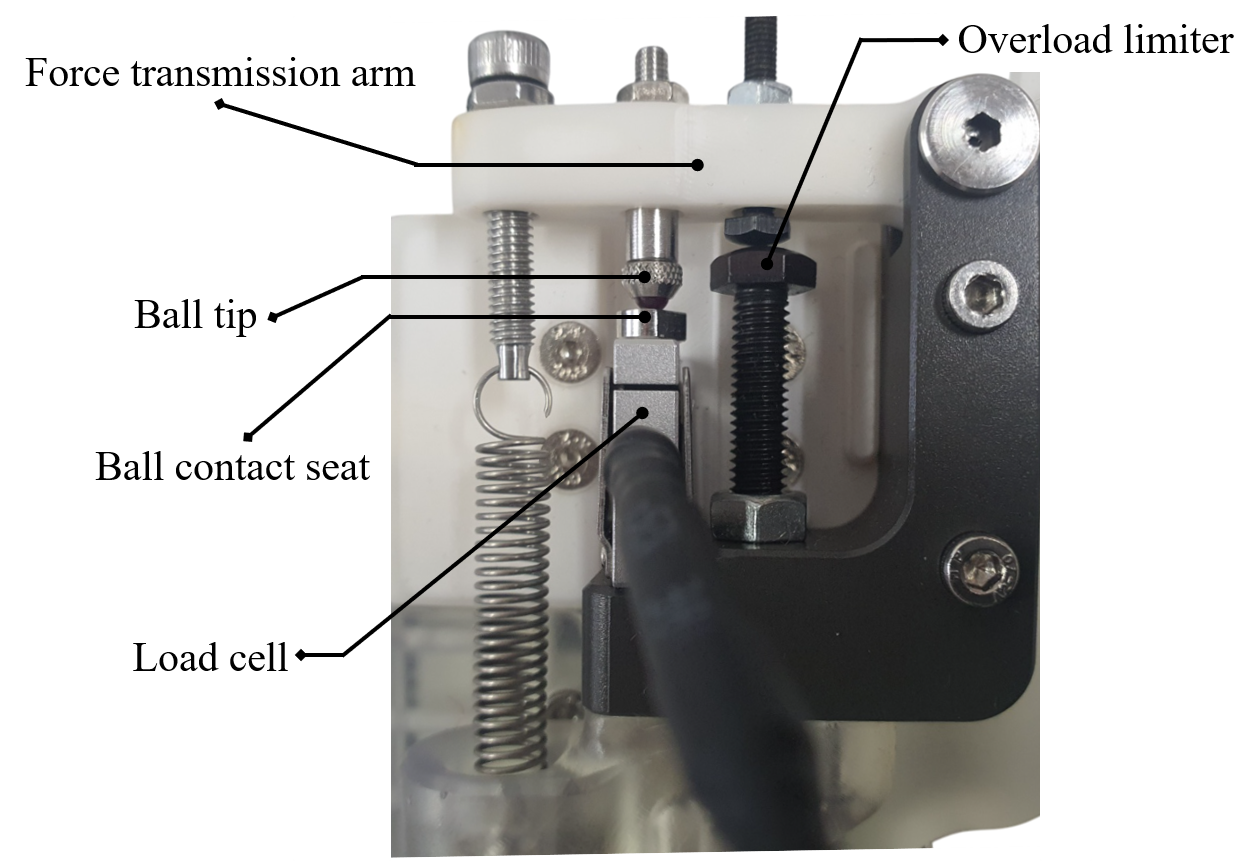} 
\vspace*{-3mm}    \caption{Implementation of the overload limiter within the sensing part to ensure clinical safety. The overload limiter prevents potential damage to the load cell by restricting excessive forces when the endoscopic robotic system interacts with human tissue, such as the kidney or the inner wall of the colon.}
   \label{fig:protecting}
\end{figure}

\subsubsection{Overload Protection Part}\label{sec:method_md_protection} 
% As mentioned in Section \ref{sec:method_requirements_safety}, to ensure clinical safety, a mechanism is needed to prevent damage to the load cell inside the sensing part due to overload. As shown in Fig. \ref{fig:protecting}, we addressed the aforementioned issue by incorporating an overload limiter within the sensing part. The overload limiter protects the load cell by limiting excessive forces and preventing potential damage under overload conditions. For example, when the endoscopic robotic system comes into contact with human kidney tissue or the inner wall of the colon, a reaction force is generated and transmitted through the insertion tube to the insertion tube holder. If excessive force is applied, the overload limiter activates to prevent potential damage to the load cell and the device (Fig. \ref{fig:sensing}(b)).
As mentioned in Section \ref{sec:method_requirements_safety}, a protective mechanism is required to prevent damage to the load cell inside the sensing part due to overload, thereby ensuring clinical safety. In general, even if moderate reaction forces occur during contact with the inner walls of the colon or kidney, it is unlikely that the load cell will experience a direct overload. In contrast, accidental impacts to the sensor section may occur when the user mounts or removes the endoscope, potentially generating a sudden large load that could damage the load cell. This may lead to inaccurate or distorted force measurements during surgery, which can undermine both the reliability of the device and the safety of the procedure.

To address this issue, we integrated an overload limiter into the sensing part, as shown in Fig. \ref{fig:protecting}. This component is designed to protect the load cell from impacts or external loads that may occur during mounting, and it can also respond in rare cases where overload is caused by abnormal collisions between the endoscopic robot and internal organ tissue. The mechanism by which the limiter interrupts force transmission under such overload conditions is illustrated in Fig. \ref{fig:sensing}(b).

\begin{figure*}[t]
    \centering
    \includegraphics[width=1.0\textwidth]{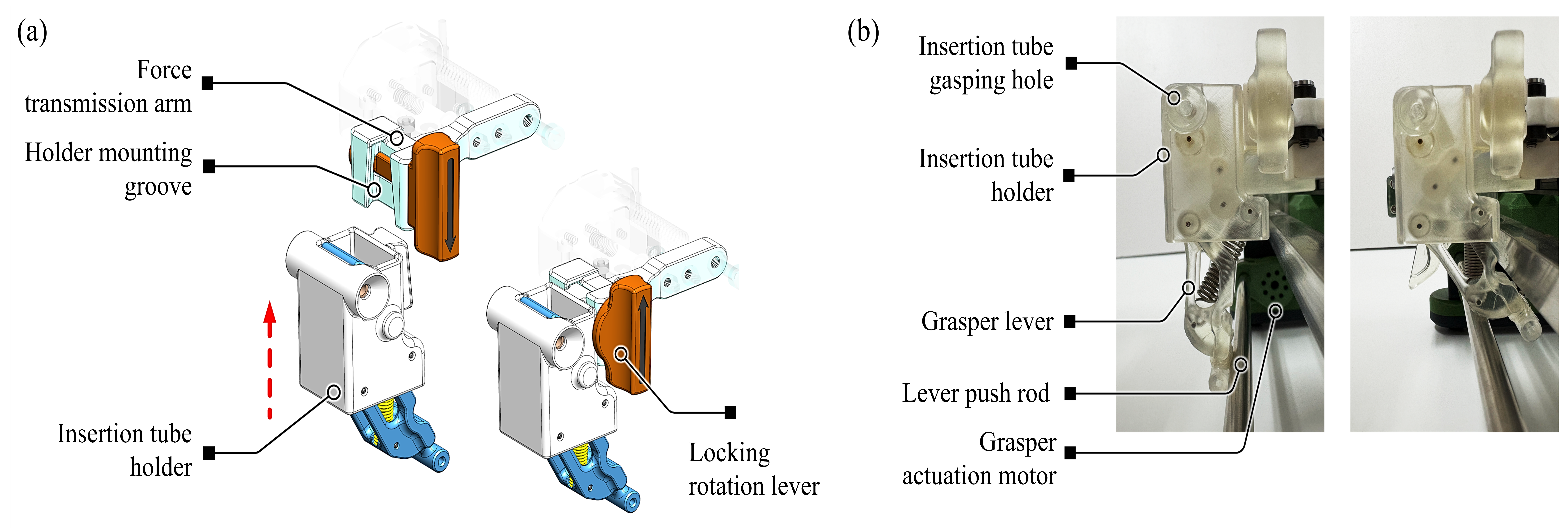}
    \caption{(a) Attachment procedure of the disposable insertion tube holder, designed to maintain sterilization by utilizing a sterile drape and minimizing cross-contamination risks; (b) bistable mechanism of the gripper part, ensuring stable holding of the insertion tube by maintaining the grasped state without continuous actuation.}
    \label{fig:gripper}
\end{figure*}

\begin{figure*}[t]
    \centering
    \includegraphics[width=0.78\textwidth]{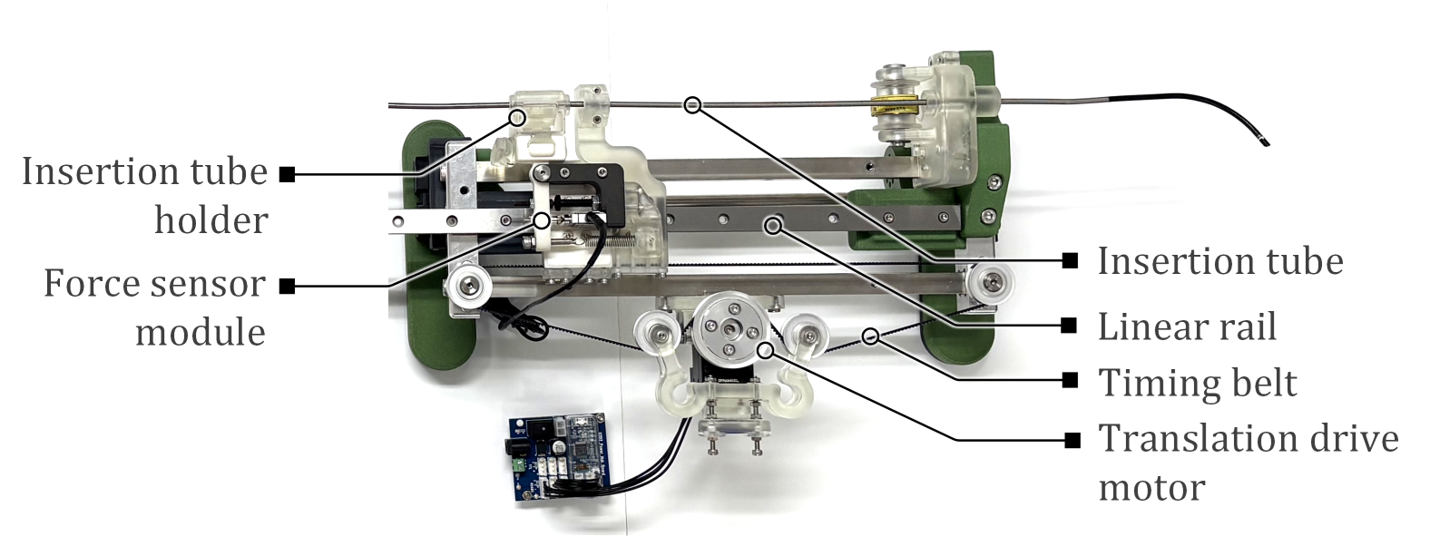}
    \caption{Operational principle of the insertion tube's forward and backward translation, along with the grasping and releasing mechanism. The system mechanically replicates the clinical procedure by sequentially grasping, advancing, and releasing the insertion tube. A servo motor and timing belt drive the linear transport part, enabling precise movement of the endoscopic robotic system while simultaneously performing axial force measurement.}
    \label{fig:transport}
\end{figure*}

\subsubsection{Gripper Part}\label{sec:method_md_gripper} 
% The EndoForce device proposed in this study, as explained in Section \ref{sec:method_md_sensor}, measures the axial force through a load cell connected to the gripper mechanism that holds the insertion tube of the endoscopic robotic system. Fig. \ref{fig:gripper} shows the overall mechanical structure of the gripper part. As mentioned in Section \ref{sec:method_requirements_sterilization}, since the insertion tube directly contacts the patient's biological tissue, the gripper part holding it should consider sterilization issues. If the gripper part and the load cell in the sensing part are designed as a single part, there may be limitations in sterilization and potential effects on sensing performance. Therefore, we designed the gripper part to be detachable, allowing for the replacement of disposable components.
The EndoForce device proposed in this study, as described in Section \ref{sec:method_md_sensor}, measures axial force through a load cell connected to a gripper mechanism that holds the insertion tube of the endoscopic robotic system. Fig. \ref{fig:gripper} illustrates the overall mechanical structure of the gripper part. 

As noted in Section \ref{sec:method_requirements_sterilization}, since the insertion tube comes into direct contact with the patient’s biological tissue, sterilization must be carefully considered in the design of the gripper part. If the gripper and the load cell in the sensing part are designed as an integrated unit, it may not only complicate sterilization but also risk degrading sensor performance during the sterilization process. Therefore, in this study, the gripper part was designed as a detachable component, allowing for easy replacement with disposable parts.

Fig. \ref{fig:gripper}(a) illustrates the attachment procedure of a disposable insertion tube holder, designed to meet sterilization requirements during endoscopic procedures by utilizing a sterile drape. With this structure, the insertion tube holder used with the drape can be easily detached and disposed of after procedures, thereby clinically reducing cross-contamination risks and effectively maintaining sterilization conditions. The disposable insertion tube holder is mounted by pushing it vertically into the holder mounting groove. This groove has a wedge-shaped design that securely fastens the insertion tube holder to the force transmission arm, preventing looseness or rattling and ensuring precise force transmission. Once properly positioned, the insertion tube holder is firmly locked in place by rotating the locking lever upward.

% The gripper part should address not only sterilization issues but also ensure the stable holding of the insertion tube. We stabilized the holding of the insertion tube by incorporating a bistable mechanism into the gripper part (Fig. \ref{fig:gripper}(b)). When the lever push rod rotates counterclockwise, it pushes the grasper lever outward, causing the insertion tube grasper to release the insertion tube. Conversely, when the lever push rod rotates clockwise, it pulls the grasper lever inward, allowing the insertion tube grasper to securely grasp the insertion tube. Moreover, the bistable mechanism ensures that the grasper remains in the grasped state even without continuous actuation of the lever push rod.

The gripper part should address not only sterilization requirements but also the ability to securely hold the insertion tube. In this study, we designed the gripper part with a bistable mechanism to ensure stable gripping of the insertion tube (Fig. \ref{fig:gripper}(b)). When the lever push rod rotates counterclockwise, the gripper lever moves outward, causing the gripper to release and the insertion tube to be disengaged. Conversely, when the lever push rod rotates clockwise, the gripper lever is pulled inward, allowing the gripper to firmly hold the insertion tube. In addition, a bistable mechanism similar to that used in a vice plier is applied between the gripper and the gripper lever. This mechanism allows the gripper to maintain a securely locked state even without continuous actuation of the lever push rod.

\subsubsection{Linear Transport Part}\label{sec:method_md_linear} 
Fig. \ref{fig:transport} illustrates the operational principle of the insertion tube's forward and backward translation, as well as the grasping and releasing mechanism. This system is designed to mechanically replicate the typical clinical procedure, where an operator manually grasps, advances, and retracts the insertion tube in a repetitive sequence. The insertion tube holder moves forward while securely grasping the tube, then releases it upon reaching the endpoint before returning to its initial position to repeat the process, enabling continuous advancement.

To automate this motion, the linear transport part is implemented using a servo motor (Dynamixel XM-430-W210-R, ROBOTIS) and a timing belt. The gripper and sensing parts are mounted on this linear drive system, enabling precise forward and backward control of the endoscopic robotic system while simultaneously performing real-time axial force measurement.

% As shown in Fig. \ref{fig:gripper}(c), the linear transport part is designed to automate the forward and backward movement of the endoscopic robotic system. This part is designed using a servo motor (Dynamixel XM-430-W210-R, ROBOTIS) and a timing belt. The gripper part and sensing part are mounted on the linear motion section of the timing belt to enable precise forward and backward movement. This allows the proposed EndoForce to simultaneously perform precise endoscope movement and axial force measurement.

\section{Experimental Design}\label{sec:experimental_design}
\begin{figure*}[t]
    \centering
    \includegraphics[width=0.9\textwidth]{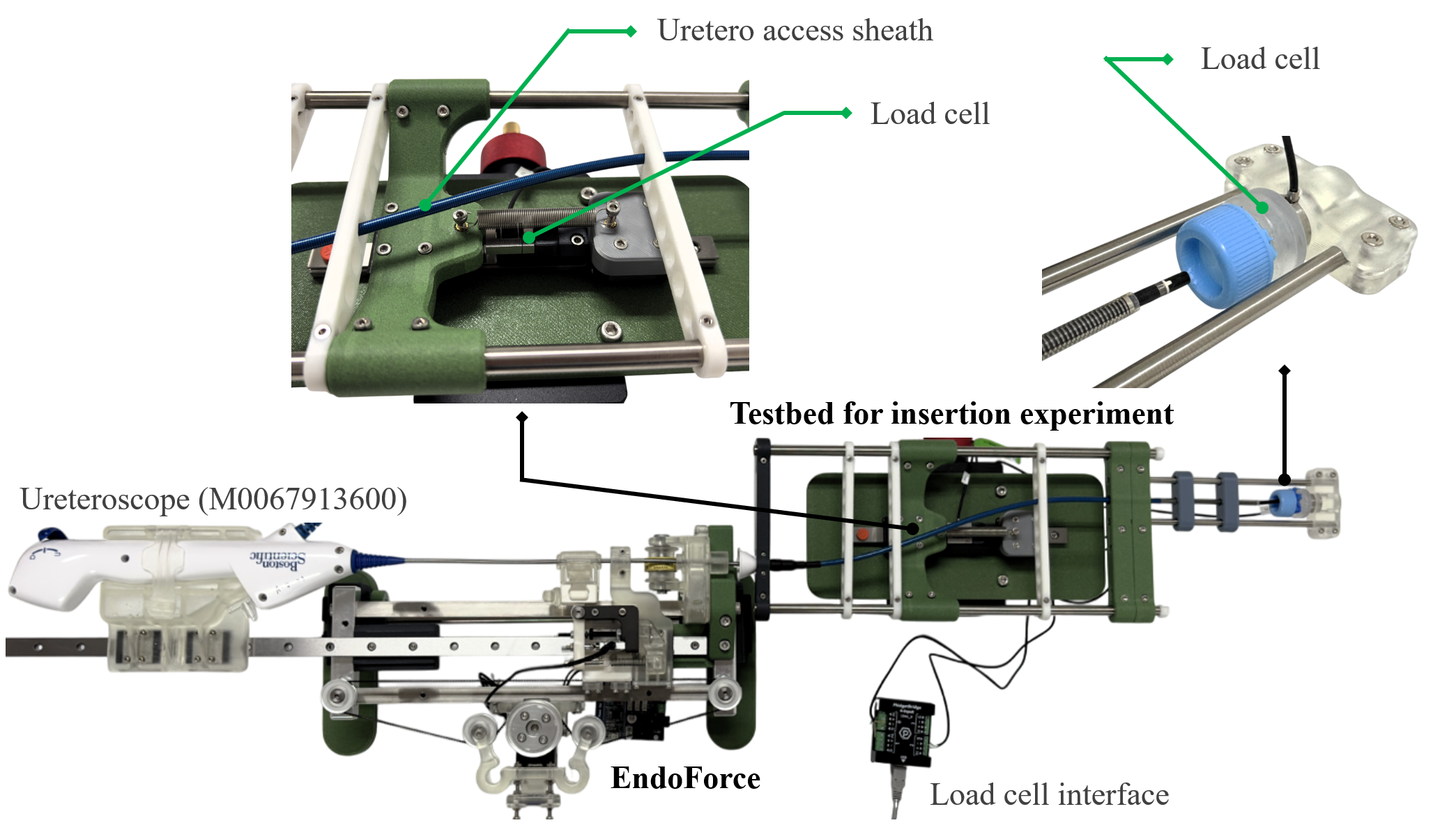}
    \caption{Experimental setup for validating the proposed force feedback device. The setup includes a commercial ureteroscope, the EndoForce, and a testbed simulating the human ureter using a 4 mm inner diameter access sheath. Straight and curved insertion pathways were created by adjusting the sheath assembly. Two load cells in the testbed measure frictional force during insertion and collision force at the distal end. The axial force measured by EndoForce is compared with the sum of these forces to evaluate its accuracy.}
    \label{fig:experimental_setup}
\end{figure*}

The main purpose of these experiments was to verify whether the axial force generated from the endoscopic robotic system during insertion into the human body is accurately measured by EndoForce, which is set up externally. Fig. \ref{fig:experimental_setup} provides an overview of the experimental setup environment. The overall experimental setup consists of three main components: a commercial ureteroscope (M0067913600, Boston Scientific), the proposed EndoForce system, and a testbed for ureteroscope insertion. 

As mentioned earlier in Section \ref{sec:introduction}, measuring the axial force from the scope is necessary not only for ureteroscopy but also for endoscopy used in GI examinations. To validate the effectiveness of the proposed EndoForce in this study, two systems can be considered: GI endoscope and ureteroscope. Considering validation in a simpler and more compact experimental setup, this study uses a commercial ureteroscope, which has a smaller overall size.

The design of the testbed for ureteroscope insertion considered the following characteristics of the actual clinical environment:
\begin{itemize}
    \item Considering the actual size of the human ureter, this study implemented an insertion pathway using a ureteral access sheath with an inner diameter of 4 mm, which is commonly used in clinical practice.
    \item Considering that the human ureter includes both straight and curved pathways, the testbed was designed to generate various insertion pathways within a single setup by simply modifying the assembly method of the ureteral access sheath.
    \item When a medical doctor manually inserts a ureteroscope into the ureter, the gripping hand can perceive force in the following two situations: (1) the force generated by friction between the ureteroscope and the ureter during the insertion process; (2) the force that occurs when the distal end of the ureteroscope collides with the ureteral wall. To measure the forces occurring in both situations mentioned above, two load cells were incorporated into the testbed used in this study. The first load cell (DYLY-108, CALT) was installed beneath the plate where the ureteral access sheath was mounted, measuring the force generated by friction as the ureteroscope was inserted. The second load cell (DYZ-100, DECENT) was positioned at the end of the testbed to measure the force occurring when the distal end of the ureteroscope collided with the ureteral wall.
\end{itemize}

The experiments were performed three times for both the straight and curved pathways. Similar to how a medical doctor inserts it in a clinical setting, the ureteroscope was stably held by the gripper part and inserted into the ureteral access sheath, which simulates the ureter, at a constant speed through linear transport part. To verify whether the proposed EndoForce can detect both of the previously mentioned situations, the ureteroscope was inserted into the testbed until a force exceeding a certain threshold was detected by the load cell mounted at its end. 

The measurements from the load cell mounted on the sensing part of the proposed EndoForce and the two load cells installed on the testbed were all processed through the load cell interface (PhidgetBridge 4-Input, Phidgets). All data output from the load cells was filtered using a moving average filter. As previously explained, the hand of the medical doctor gripping the ureteroscope experiences resistance both during insertion and upon collision. Given that the force measured by the load cell in the sensing part should ideally be equal to the sum of the force values from two load cells installed in the testbed, this experiment defines the Root Mean Squared Error (RMSE) between the two values as the evaluation criterion.

% The human urinary tract includes both straight and curved pathways \cite{polter2015risk, kim2021sigmoid, thakur2020trapped, huynh2017retained, gadzhiev2017valve}.
% The proposed EndoForce system has been designed with this consideration in mind, reflecting this requirement in its overall mechanical structure.

\section{Results}\label{sec:results}
\begin{figure*}[t]
    \centering
    \includegraphics[width=1.0\textwidth]{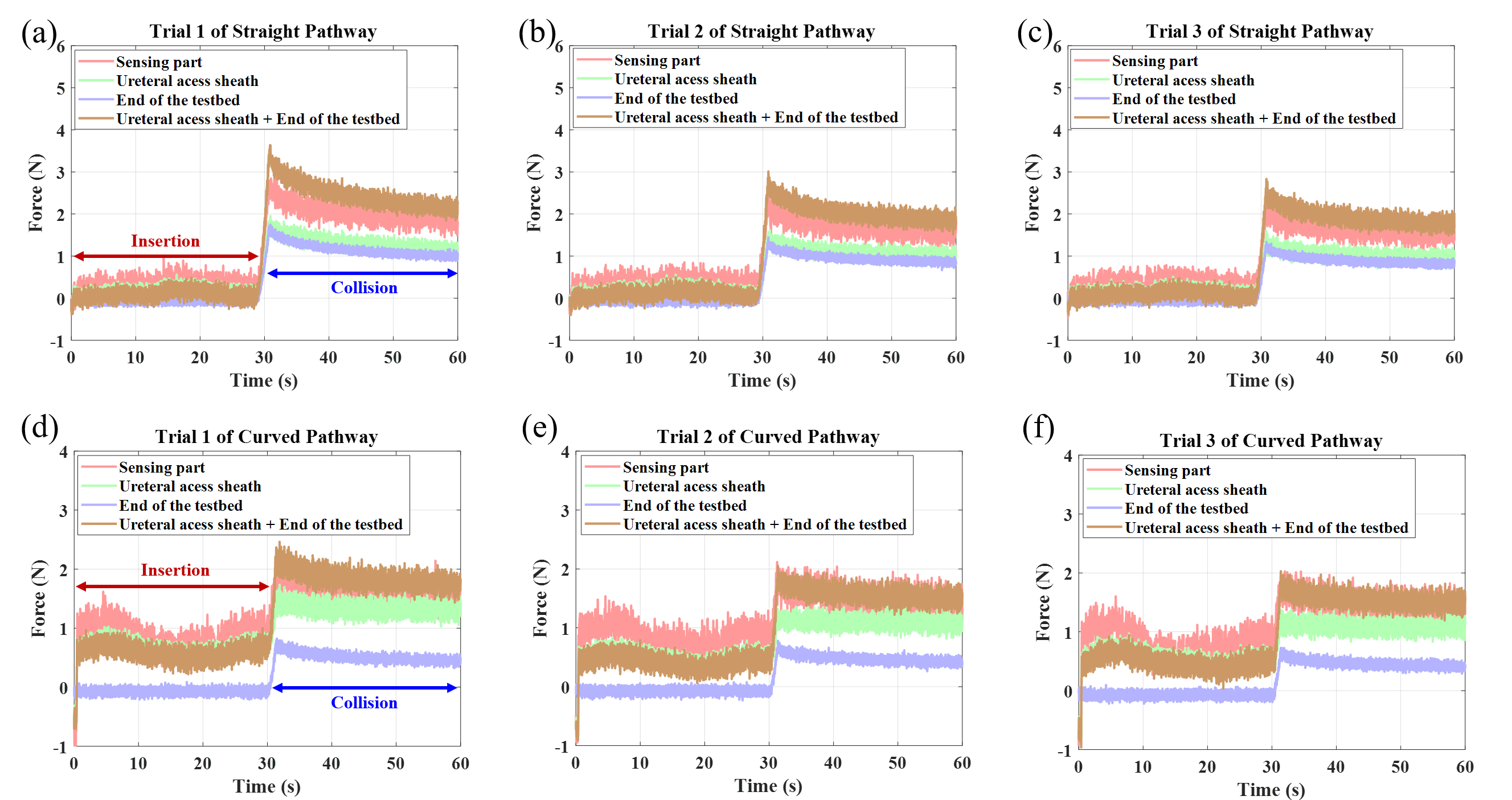}
    \caption{(a)-(c) Experimental results of ureteroscope insertion along a straight pathway. The light red data represent the force values measured by the load cell on the proposed EndoForce. The light green and light blue data correspond to the force values measured by the load cell on the plate with the ureteral access sheath and the load cell at the end of the testbed, respectively; and (d)-(f) the results of ureteroscope insertion along a curved pathway, using the same color scheme as the straight pathway experiment.}
    \label{fig:experimental_results}
\end{figure*}

The results of the physical experiment sensing the axial force transmitted from the endoscopic robotic system using the proposed EndoForce in this study are shown in Fig. \ref{fig:experimental_results}.

Fig. \ref{fig:experimental_results}(a)-(c) shows the results from the experiment where the ureteroscope was inserted along a straight pathway. The data shown in light red represent the force values measured by the load cell mounted on the proposed EndoForce in this study. The data shown in light green and light blue represent the force values measured by the load cell installed on the plate with the ureteral access sheath and the load cell installed at the end of the testbed, respectively. Considering the evaluation criteria defined in Section \ref{sec:experimental_design}, the sum of the force values measured by the two load cells mounted on the testbed is represented in dark orange. As shown in Fig. \ref{fig:experimental_results}(a)-(c), the entire experiment lasted approximately 60 sec, with the first 30 sec representing the insertion process and the latter 30 sec showing the results after the collision with the end of the testbed. Throughout the entire experiment, the RMSE between the force measured by EndoForce and the sum of the values from the two load cells mounted on the testbed was calculated to be approximately 0.43 N.

% During the insertion process, the average force measured by EndoForce was approximately 0.34 N, while the average force measured after the collision was around 1.80 N.

Fig. \ref{fig:experimental_results}(d)-(f) summarize the results from the experiment where the ureteroscope was inserted along a curved pathway. The experimental results for the curved pathway were also visualized using the same colors as those used for the straight pathway experiment. The entire experiment was performed for a total of 60 sec, similar to the straight pathway insertion experiment. The first 30 sec correspond to the insertion process along the curved pathway, while the latter 30 sec represent the results after the collision with the end of the testbed. Throughout the entire experiment, the RMSE between the force measured by EndoForce and the sum of the values from the two load cells mounted on the testbed was calculated to be approximately 0.39 N.

% During the insertion process, the average force measured by EndoForce was approximately 0.83 N, while the average force measured after the collision was around 1.53 N.
\section{Discussion}\label{sec:discussion}
In this study, we propose a system named EndoForce, which can measure axial force generated when a flexible device, such as an endoscopic robotic system, is inserted into the human body. The proposed system not only accurately measures axial force but also takes into account reliability, cost-effectiveness, safety, and sterilization issues in its overall design for practical clinical application.

As explained in Section \ref{sec:experimental_design}, this study validated the effectiveness of the proposed EndoForce through physical experiments on a robotic system that includes a ureteroscope. Although a ureteroscope with a relatively smaller diameter and shorter length was used in this study to establish a simpler and more compact experimental environment, EndoForce is expected to be applicable to various commercially available endoscopes and endoscopic robotic systems if certain mechanisms are improved. In this study, a gripper part utilizing a trigger mechanism was proposed to stably hold a device with a specific diameter. However, future research will focus on developing a mechanism that can flexibly adapt to changes in the diameter of the mounted device. Additionally, the forward and backward motion of the ureteroscope was controlled through a linear transport part consisting of a linear guide and timing belt. The overall length of the system, consisting of the linear guide and timing belt, increases depending on the maximum depth to which the scope or robot needs to be inserted. Although this study focused on verifying the effectiveness of accurate axial force measurement, this issue was not critical. However, it is believed that a more efficient mechanism for forward and backward motion will be necessary to apply it to GI endoscopes or endoscopic robotic systems, which can be up to 2 meters in length.

Fig. \ref{fig:experimental_results} shows the overall results of the physical experiments performed in this study. As mentioned in Section \ref{sec:results}, the proposed EndoForce system was shown to accurately measure the axial force generated during the insertion of the ureteroscope (The RMSE was measured to be approximately 0.43 N along the straight pathway, and approximately 0.39 N along the curved pathway.). Despite processing the values from the load cell mounted on the sensing part using a moving average filter, there was still an average standard deviation of approximately 0.45 N (See all the data represented in lite red in Fig. \ref{fig:experimental_results}). In the future, applying AI technologies such as Long Short-Term Memory (LSTM) \cite{jang2023spectral}, which learns the temporal dependencies in time-series data to capture long-term patterns and reduce noise, or Generative Adversarial Networks (GAN) \cite{xiang2018speech}, which can transform noisy data into cleaner data through the competition between the generator and discriminator, is expected to improve the noise in the load cell data and yield more meaningful results.

Although the EndoForce proposed in this study focuses on measuring axial force, future research should ultimately integrate a haptic feedback system based on the measured force values. Therefore, when remotely controlling the endoscopic robotic system, a motor or vibration device will be connected to the master device to transmit the force values measured by EndoForce to the medical doctor in real time. This haptic feedback system is expected to ensure precise manipulation and patient safety by transmitting the resistance felt by the medical doctor when manually inserting a ureteroscope or GI endoscope, even when the robot is controlled remotely. Additionally, recent advancements in artificial intelligence (AI) technology have led to active research on automating sub-tasks of surgical procedures using robotic systems in laboratory environments \cite{hwang2022automating, dharmarajan2023automating, hari2024stitch, yang2024development, ramakrishnan2024automating}. Even though automation research so far has been performed using rigid-type robotic systems like the Da Vinci system, it is expected that research combining the proposed EndoForce with a haptic feedback system could lead to meaningful results in the future automation of endoscopic robotic systems.

\section{Concluding Remarks}\label{sec:conclusion}
This study proposes EndoForce, a new device inspired by the actual insertion motion of medical doctors during conventional ureteroscopy and GI endoscopy. It accurately measures the axial force from the endoscopic robotic system, providing intuitive force perception. The device is designed for easy attachment and detachment of the sterile cover, ensuring compatibility with real clinical environments. Additionally, the use of a commercial load cell makes the device cost-effective, with a simplified design and application.

In future work, research will focus on developing a more adaptable gripper mechanism that can accommodate variations in device diameter and improving the efficiency of the forward and backward motion mechanism for application in longer endoscopic robotic systems. Additionally, AI-based noise reduction techniques such as LSTM and GAN will be explored to enhance the accuracy of load cell data processing. Lastly, research will be performd on integrating EndoForce with a haptic feedback system to enable real time force transmission to medical doctors, ensuring precise manipulation and patient safety, as well as exploring its potential application in the future automation of endoscopic robotic systems.

\appendices
\bibliographystyle{IEEEtran}
\bibliography{bibliography}

\section{Author Information} % should remove this section 
\begin{IEEEbiography}[{\includegraphics[width=1in,height=1.25in,clip,keepaspectratio]{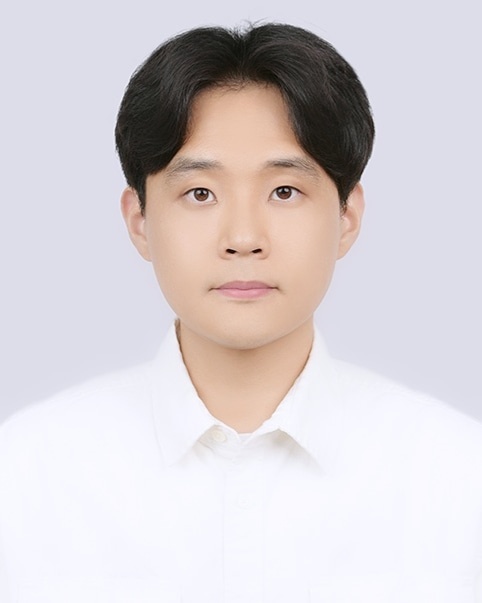}}]{Hansoul Kim} received the B.S. degree from the Division of Robotics, Kwangwoon University, Seoul, Republic of Korea, in 2017, the M.S. degree from the Robotics Program, Korea Advanced Institute of Science and Technology (KAIST), Daejeon, Republic of Korea, in 2019, and the Ph.D. degree from the Department of Mechanical Engineering, KAIST, in 2022.

From 2022 to 2023, he served as a Post-doctoral researcher at the Walker Department of Mechanical Engineering and Texas Robotics, the University of Texas at Austin, TX, USA. From 2023 to 2024, he served as as a Post-doctoral researcher in the Department of Electrical Engineering and Computer Sciences at the University of California, Berkeley, CA, USA. He 000is currently an Assistant Professor with the Department of Mechanical Engineering at the Myongji University, Yongin, Republic of Korea. His research interests primarily focus on surgical robotics, AI-assisted robot surgery, robot task automation, soft robotics, and mechanism design.
\end{IEEEbiography}

\begin{IEEEbiography}[{\includegraphics[width=1in,height=1.25in,clip,keepaspectratio]{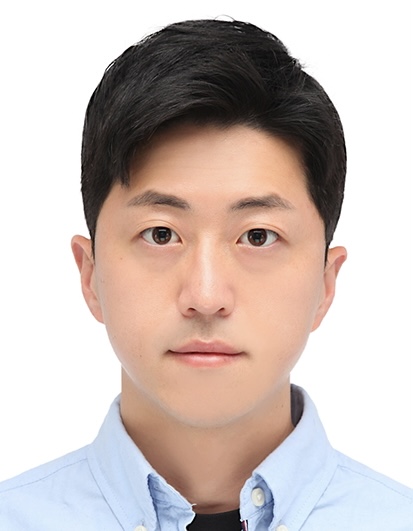}}]{Dong-Ho Lee} received the B.S. degree in the Department of Mechanical System Design Engineering from Seoul National University of Science and Technology, Seoul, Republic of Korea, in 2014, and M.S. and Ph.D. degrees in robotics program from the Korea Advanced Institute of Science and Technology (KAIST), Daejeon, Republic of Korea, in 2016, and 2021, respectively. 

He is currently CTO (Chief Technology Officer) with ROEN Surgical Inc., Daejeon, Republic of Korea. His research interests include surgical robotics, mechanism design, and AI with its applications for surgical robots.
\end{IEEEbiography}

\begin{IEEEbiography}[{\includegraphics[width=1in,height=1.25in,clip,keepaspectratio]{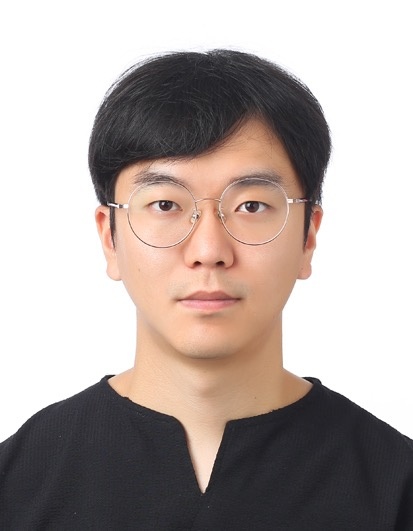}}]{Dukyoo Kong} received the B.S. degree in mechanical engineering from Korea Advanced Institute of Science and Technology (KAIST), Daejeon, Republic of Korea, in 2014, and the M.S. and Ph.D. degrees in the robotics program from KAIST in 2016 and 2022, respectively. He is currently the Customer Service Team Manager at ROEN Surgical Inc., Daejeon, Republic of Korea. His research interests include surgical robotics and force feedback for surgical robots.
\end{IEEEbiography}

\begin{IEEEbiography}[{\includegraphics[width=1in,height=1.25in,clip,keepaspectratio]{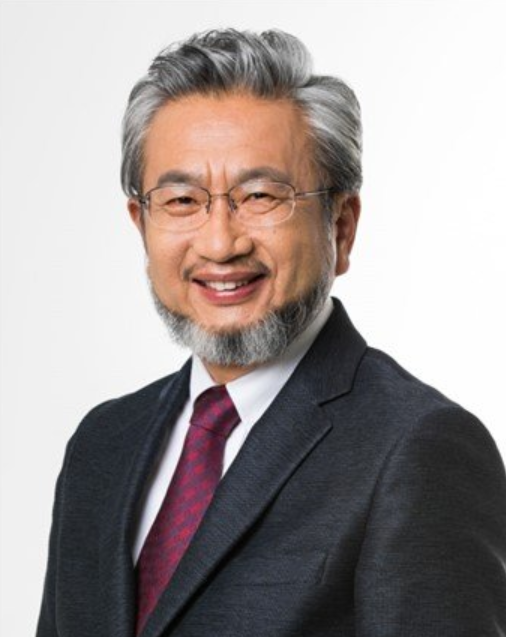}}]{Dong-Soo Kwon} received the B.S. degree in mechanical engineering from Seoul National University, Seoul, Republic of Korea, in 1980, the M.S. degree in mechanical engineering from Korea Advanced Institute of Science and Technology (KAIST), Daejeon, Republic of Korea, in 1982, and the Ph.D. degree in mechanical engineering from the Georgia Institute of Technology, Atlanta, GA, USA, in 1991. From 1991 to 1995, he was a Research Staff Member at Oak Ridge National Laboratory, Oak Ridge, TN, USA. He joined the Department of Mechanical Engineering, KAIST, in 1995, where he served as Professor and is currently an Emeritus Professor. He is also the CEO of ROEN Surgical. His research interests include medical robotics, haptic interfaces, telerobotics, human–robot interaction, surgical robotic systems, and parallel manipulators. He is a member of IEEE, KSME, and ICASE.
\end{IEEEbiography}

\begin{IEEEbiography}[{\includegraphics[width=1in,height=1.25in,clip,keepaspectratio]{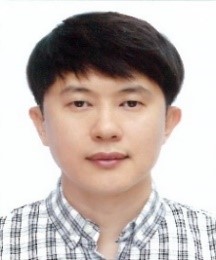}}]{Byungsik Cheon} received his Bachelor of Engineering from Korea University of Technology and Education in 2007, his Master of Engineering in Robotics from Daegu Gyeongbuk Institute of Science and Technology (DGIST) in 2013, and his Ph.D. from the Robotics Program at the Korea Advanced Institute of Science and Technology (KAIST) in the Republic of Korea in 2022. He is currently an Assistant Professor in the School of Mechatronics Engineering at the same university where he earned his bachelor's degree, with his primary research focus on surgical robots.
\end{IEEEbiography}
\EOD

% \bibliographystyle{IEEEtran}
% \bibliography{bibliography}
\end{document}